# Prominent Attribute Modification using Attribute Dependent Generative Adversarial Network

Naeem Ul Islam, Sungmin Lee, and Jaebyung Park*

*Abstract*— Modifying the facial images with desired attributes is important, though challenging tasks in computer vision, where it aims to modify single or multiple attributes of the face image. Some of the existing methods are either based on attribute independent approaches where the modification is done in the latent representation or attribute dependent approaches. The attribute independent methods are limited in performance as they require the desired paired data for changing the desired attributes. Secondly, the attribute independent constraint may result in the loss of information and, hence, fail in generating the required attributes in the face image. In contrast, the attribute dependent approaches are effective as these approaches are capable of modifying the required features along with preserving the information in the given image. However, attribute dependent approaches are sensitive and require a careful model design in generating high-quality results. To address this problem, we propose an attribute dependent face modification approach. The proposed approach is based on two generators and two discriminators that utilize the binary as well as the real representation of the attributes and, in return, generate high-quality attribute modification results. Experiments on the CelebA dataset show that our method effectively performs the multiple attribute editing with preserving other facial details intactly.

## I. INTRODUCTION

The biological identity of a human face has been extensively studied in computer vision and robotics ranging from face identification and detection [1,2,3,4] to face attribute modification [5,6,7]. For example, smart devices such as mobile phones, service robots and surveillance cameras capture the images of people. However, proper investigation is required for an accurate identification of a person by modifying his face attributes accordingly. Furthermore the success of the above-mentioned approaches has been possible by three facts. (i) Numerous publically available training data with labels, (ii) High computational capabilities in terms of GPUs, and (iii) open-source libraries. Although, under the availability of the above-mentioned resources, a huge amount of work has been done in face identification, detection and attribute modification, more attention is required in terms of the prominent attribute modification while keeping the rest of the features intact along with proper interpolation capabilities. Furthermore, the facial attribute modification task is more challenging as compared to face recognition and detection, where it requires a careful description of the semantic aspects of the face while modifying the required attributes and keeping the face identity intact. For example, if we need to modify a specific attribute such as changing the color of the hair, we need to have semantic information about the hair and modify only that part of the image without affecting the other attributes of the face image.

Facial attribute modification is related to image editing that has been extensively studied in computer graphics in terms of different applications such as color modification [8], content modification [9] and image wrapping [10, 11]. Two kinds of approaches have been used to handle image editing problem: example-based [12, 13, 14] and model-based [11, 15]. The example-based approach searches the required attribute in the given reference image and transfers it to the target image. This enables the image editing in various ways depending on the available reference image [12, 13, 14]. However, it requires the reference image to be of the same person with the proper face alignment and the same lighting condition. The model-based approach first builds the model of the required face and then modifies the image accordingly [11, 15]. Although these approaches are successful in particular attribute modification, they are task-specific and cannot be applied to arbitrary attribute modification problems.

Recent advances in deep neural networks such as generative adversarial networks (GANs) [16] and variational autoencoders (VAEs) [17] pave the way to several face attribute modification approaches [18, 19] and distortion removal approaches in real images [20]. Both GANs and VAEs are powerful models and are capable of generating images. The GAN generates more realistic images than the VAE, but it can't encode images since it uses random noise as an input. In contrast, the VAE has the capability of encoding the image to its corresponding latent representation, but the generated image is blurry as compared to the GAN. The combination of the GAN and the autoencoder provides a powerful tool for image attribute editing. IcGAN [21] and cGAN [22] use the combination of the GAN and the autoencoder for editing the attributes of an image by modifying the latent representation to include the information of the expected attribute and then decode the modified image. GeneGAN [5] learns the object transfiguration from two unpaired sets of images: one set of images with specific attributes while the other set of images doesn't have those attributes. There is a mild constraint here that the objects are located approximately at the same place. For example, the training data can be one set of reference face images with eyeglasses and another set of images without eyeglasses, where both of them are spatially aligned by face landmarks. DNA-GAN [23] can be considered as an extension of the

*Research supported by Basic Science Research Programs, (No. 2019R1A6A1A0931717) and (NRF-2018R1D1A1B07049270), through the National Research Foundation of Korea (NRF) funded by the Ministry of Education.

Naeem Ul Islam is with Core Research Institute of Intelligent Robots, Jeonbuk National University, Jeonju, Korea (e-mail: naeem@jbnu.ac.kr).

Sungmin Lee is with Core Research Institute of Intelligent Robots and Division of Electronics Engineering, Jeonbuk National University, Jeonju, Korea (e-mail: leesungmin @jbnu.ac.kr)

Jaebyung Park* (corresponding author) is with Core Research Institute of Intelligent Robots and Division of Electronics Engineering, Jeonbuk National University, Jeonju, Korea (e-mail: jbpark@jbnu.ac.kr).

GeneGAN, which makes "crossbreed" images by swapping the latent representation of the corresponding attributes between the given pair of images. These methods, however, are useful in image editing tasks, but they require different models for different attributes that are not required in real-world applications. The proposed work is based on the attribute dependent approach, where it applies an attribute classification constraint to the generated images. AttGAN [6] has the capability of changing the required attribute while keeping the other details unchanged. However, the modified attributes are not prominent even though the face identity is well preserved. In contrast, the proposed approach focuses on the prominent modification of the attributes while keeping the rest of the attributes unchanged. Recently, RelGAN has been proposed in [7], where they can effectively modify the attributes simultaneously with an additional capability of interpolation. However, they obtained the relative attributes by taking the difference between the given and the predefined target attributes in their proposed approach. In contrast, the proposed approach doesn't take the predefined attributes but utilizes the already available target attributes along with their mean representation.

## II. PROPOSED APPROACH

The proposed attribute modification approach in Fig. 1 is attribute dependent. That is, it applies attribute classification constraint to the generated images. The proposed approach consists of two generators and two discriminators. Both the generators take as input the face image that needs to be modified. The encoders of both the generators share the same parameters as they take the same input image and project the input image to the latent representation $z$. The decoder of the first generator $G1_{dec}$ takes as input the latent representation of the given input image along with the required attribute $b$ as well as the given attribute vector $a$. It then decodes them respectively, to the modified image $x^{b'}$ and reconstructed image $x^{a'}$. Similarly, the decoder $G2_{dec}$ decodes the combination of the latent representation of the given input image and the mean attribute vector. In terms of the discriminators, the first discriminator $Disc1$ takes either the real image $x^a$ or fake image $x^{b'}$ generated by $G1_{dec}$ and maps it to the attribute vectors $a$ or $b$ along with real and fake classification labels. In return, it guides the first generator to generate the modified image. Similarly, $Disc2$ takes the real image $x^a$ or fake image $x^{c'}$ generated by $G2_{dec}$ and maps it to the corresponding attribute vectors $a$ or $b$ along with real and fake classification labels. In return, it guides the second generator.

## III. TRAINING

The purpose of the proposed attribute modification approach is to modify multiple attributes simultaneously while utilizing the attribute information in the input data. The generator takes the required image as an input, where we wish to modify some of its contents by decoding a different combination of the attributes. In this approach, we selected thirteen attributes that are to be modified. These attributes include baldness, bangs, black hair, blond hair, brown hair, bushy eyebrows, eyeglasses, gender, mouth open/closed, mustaches, no beard, pale skin, and young. For example, if the black hair or mustaches are the desired attributes in the input testing sample, this approach will make the hair black and put mustaches on the face of the given image. During the training, the attributes of each image are concatenated in three different configurations. In the first configuration, the original attributes are concatenated with the latent representation of the input image, and then the combination of the attributes and the latent representation are decoded back to the original image. The original attribute objective along with the GAN objective is used to train the first and second discriminators, $Disc1$ and $Disc2$, of the network. In the second configuration, the input attributes are first shuffled and then concatenated with the latent representation of the input image. Next, the concatenated pair is decoded back to the modified image using $G1_{dec}$. In the third configuration, the mean value of the input attributes and the shuffled attributes are concatenated with the latent representation of the input image. Next, the concatenated pair is decoded back to the modified image using $G2_{dec}$. The attribute objective along with reconstruction and GAN objectives is used to tune the parameters of both the generators of the network.

The detailed explanation and stepwise training procedure are as follows.

For a given face input image $x^a$, its attributes $a$ are defined as follows.

$$a = [a_1, a_2, \ldots, a_n] \quad (1)$$

The encoder part of the attribute modification network takes $x^a$ as an input and transforms it into its corresponding latent representation:

$$z = Enc_{Gen}(x^a) \quad (2)$$

During the first configuration, the decoder $G1_{dec}$ translates the latent representation $z$ along with its attribute vector $a$ to $x^{a'}$ as:

$$x^{a'} = G1_{dec}(z, a) \quad (3)$$

During the second configuration, the attribute vector is first modified as:

$$b = f(a) \quad (4)$$

The decoder $G1_{dec}$ then translates the latent representation $z$ along with the required attribute vector $b$ to $x^{b'}$:

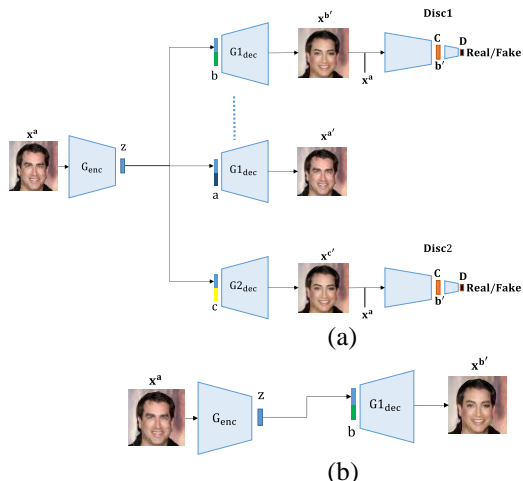

Fig. 1. Face attribute modification network: a) Training, b) Testing.

$$x^{b'} = G1_{dec}(z, b) \tag{5}$$

During the third configuration, the mean attribute vector is first obtained as:

$$c = (a + f(a))/2 \tag{6}$$

The decoder $G2_{dec}$ then translates the latent representation $z$ along with the required attribute vector $c$ to $x^{c'}$.

$$x^{c'} = G2_{dec}(z, c) \tag{7}$$

The GAN objective functions for training, respectively, the discriminators, $Disc1$ and $Disc2$, are defined as follows.

$$L_{Disc1\_GAN} = \gamma_d \left( Disc1(x^a) + 1 - Disc1(x^{b'}) \right) \tag{8}$$

$$L_{Disc2\_GAN} = \gamma_d \left( Disc2(x^a) + 1 - Disc2(x^{c'}) \right) \tag{9}$$

$L_{Disc1\_GAN}$ and $L_{Disc2\_GAN}$ are the GAN losses for training both the discriminators. $Disc1(x^a)$ and $Disc2(x^a)$, respectively, represent the outputs of the discriminators, $Disc1$ and $Disc2$, for the given original input that needs to be modified. $Disc1(x^{b'})$ and $Disc2(x^{c'})$, respectively, represent the outputs of the discriminators, $Disc1$ and $Disc2$, for the given modified face image generated, respectively, by the $G1_{dec}(.)$ and $G2_{dec}(.)$. The attribute objective for preserving the remaining attributes of the face image is:

$$LDisc1_{att\_a} = -\sum_{i=1}^{n} a_i \log Disc1(x_i^a)$$

$$= -a_i \log(Disc1(x_i^a)) - (1 - a_i) \log(1 - Disc1(x_i^a)) \tag{10}$$

$LDisc1_{att\_a}$ represents the sigmoid cross-entropy loss for the given attributes, where $a_i$ represents the target attribute and $Disc1(x_i^a)$ represents the generated predicted attributes.

$$LDisc2_{att\_a} = -\sum_{i=1}^{n} a_i \log Disc2(x_i^a)$$

$$= -a_i \log(Disc2(x_i^a)) - (1 - a_i) \log(1 - Disc2(x_i^a)) \tag{11}$$

$LDisc2_{att\_a}$ represents the sigmoid cross-entropy loss for the given attributes, and $Disc2(x_i^a)$ represents the generated predicted attributes.

The overall objective for training the discriminator, $Disc1$, is defined as follows.

$$L_{Disc1} = L_{Disc1\_GAN} + \alpha LDisc1_{att\_a} \tag{12}$$

where $\alpha$ represents the control parameter for the original attribute objective. Next, the overall objective for training the discriminator, $Disc2$, is defined as:

$$L_{Disc2} = L_{Disc2\_GAN} + \alpha LDisc2_{att\_a} \tag{13}$$

Similarly, the training objective functions for the first and second generators of the network are defined as:

$$L_{G1\_GAN} = \gamma_d \left( 1 - Disc1(x^{b'}) \right) \tag{14}$$

$$L_{G2GAN} = \gamma_d \left( 1 - Disc2(x^{c'}) \right) \tag{15}$$

$L_{G1\_GAN}$ and $L_{G2\_GAN}$, respectively, are the GANs losses for training the first and second generators.

$$LDisc1_{att\_b} = -\sum_{i=1}^{n} b_i \log Disc1(x_i^{b'})$$

$$= -b_i \log \left( Disc1(x_i^{b'}) \right) - (1 - b_i) \log \left( 1 - Disc1(x_i^{b'}) \right) \tag{16}$$

$$LDisc2_{att\_c} = -\sum_{i=1}^{n} c_i \log Disc2(x_i^{c'})$$

$$= -c_i \log \left( Disc2(x_i^{c'}) \right) - (1 - c_i) \log \left( 1 - Disc2(x_i^{c'}) \right) \tag{17}$$

$LDisc1_{att\_b}$ and $LDisc2_{att\_c}$, respectively, represent the sigmoid cross-entropy losses for the modified attributes, $b_i$ and $c_i$ where $b_i$ represents the target binary attributes, $c_i$ represents the target mean attributes, and $Disc1(x_i^{b'})$ and $Disc2(x_i^{c'})$ represent the corresponding generated predicted attributes.

$$L_{recons} = \zeta(abs(x^a - x^{a'})) \tag{18}$$

$L_{recons}$ represents the absolute reconstruction objective, where the aim is to preserve the remaining attributes of the face image during the modification of the required attribute objectives. $x^a$ represents the input face image while $x^{a'}$ shows the reconstructed image with the intent to preserve the remaining features of the given face image. $\zeta$ shows the control parameter for preserving the remaining features while using the reconstruction objective.

The overall training objective for the generator part of the network is given below.

$$L_{Gen1} = L_{G1\_GAN} + \lambda LDisc1_{att\_b} + L_{recons} \tag{19}$$

$$L_{Gen2} = L_{G2\_GAN} + \lambda LDisc2_{att\_c} + L_{recons} \tag{20}$$

where $\lambda$ represents the control parameter for the required attribute objective, and $L_{Gen1}$ and $L_{Gen2}$, respectively, represent the losses of the first and second generators.

## IV. EXPERIMENTS

For training the proposed network, we used the CelebA dataset [24]. The CelebA is a large-scale face dataset, which is composed of 202599 face images. We divided the CelebA dataset into training and testing sets. The training set is composed of 182000 images, and the testing set is composed of the remaining 20599 images. After training, we evaluated the network for modifying the input images for the required attributes. The effectiveness of the proposed approach is proved based on different experiments for all the attributes.

In the first experiment, we evaluated the proposed approach in terms of the baldness of the given input image. If the given image is not bald, the proposed approach will inverse the bald attribute and in response generates the bald image. The comparative qualitative results of the proposed approach with AttGAN [6] in terms of the baldness are provided in Fig. 2. The left side of Fig. 2 shows the results from the AttGAN, whereas the right side shows the outputs of the proposed approach. From this analysis, we can observe that the AttGAN generates smoother results but this smoothness affects the quality of the modified image. In other words, the baldness effect is not distinct in the results of the AttGAN as compared to our results.

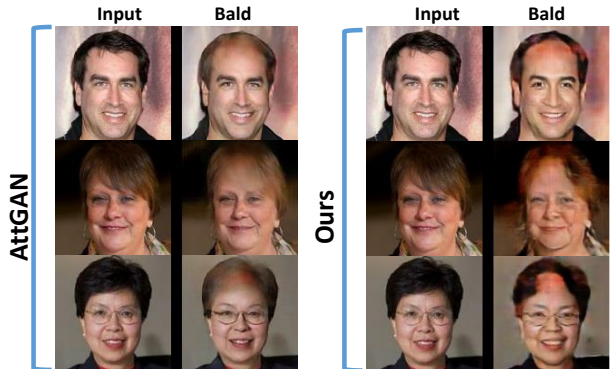

Fig. 2. Modification of attributes in face images. Left: AttGAN [6]. Right: Proposed approach. The input images with hair are transformed to the output images that are bald.

The qualitative comparative analysis in Fig. 3 shows the gender transformation. For example, if the given input image is male, the resulting generated image will be female and vice versa. Here, we can observe that the transformed images from the proposed approach show more resemblance to the required attributes as compared to the AttGAN approach.

Similarly, Fig. 4 shows the qualitative comparative analysis in terms of putting or removing the mustaches from the given face images. The left side of the figure shows the results from the AttGAN while the right side shows the results from the proposed approach. The input images without mustaches in odd columns are transformed to the output images with mustaches as shown in the even columns. From this analysis, we can observe that the AttGAN approach successfully transforms the given input images without mustaches to the images with mustaches while keeping the other features of the input images well preserved. However, the required modified features are not prominent. In contrast, the proposed approach transforms the given input images without mustaches to images with mustaches, where the required features are modified more prominently. However, the remaining attributes are slightly affected in the given input

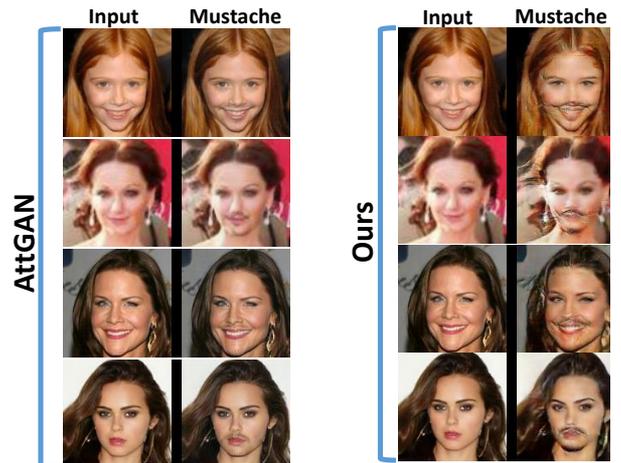

Fig. 4. Modification of attributes in face images. Left: AttGAN. Right: Proposed approach. The input images without mustaches are transformed to the output images with mustaches.

images.

In the last analysis, we performed a comparative qualitative analysis of the proposed approach with the AttGAN approach in terms of all of the thirteen attributes as shown in Fig. 5. The upper part of Fig. 5 shows the results generated by the AttGAN approach while the lower part shows the results from the proposed approach. The input images in the first column are the candidate samples that need to be modified according to the required attributes. The second column shows the reconstructed results from both approaches, whereas the rest of the columns show the modified results from bald to young. These results show that the proposed approach has the capability of modifying multiple attributes in an effective way.

## V. CONCLUSION

In this paper, we aim at modifying the multiple attributes in the given input images by utilizing the attribute dependent classification approach. The proposed approach possesses the capability of modifying the input images in an effective way as shown in the experimental section. The proposed approach has the capability of prominently modifying the given attributes while it slightly affects the other features in the input images. Furthermore, since interpolation is a desirable feature in image modification, in as our future work, we are going to extend the proposed approach with the additional capability of interpolation along with prominent attribute modifications while keeping the other features well preserved.

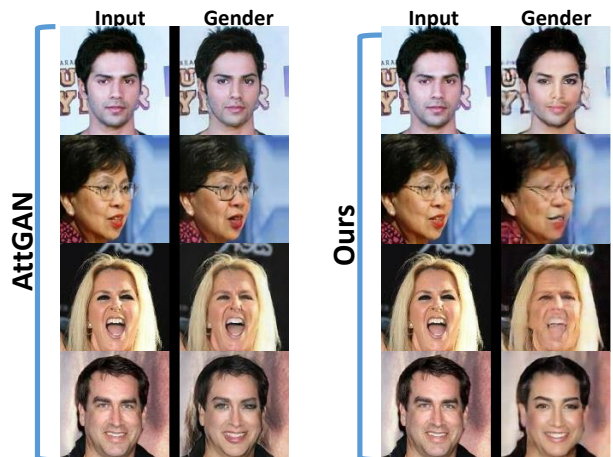

Fig. 3. Modification of attributes in face images. Left: AttGAN. Right: Proposed approach. The male input images are transformed to the female output images, and vise versa.

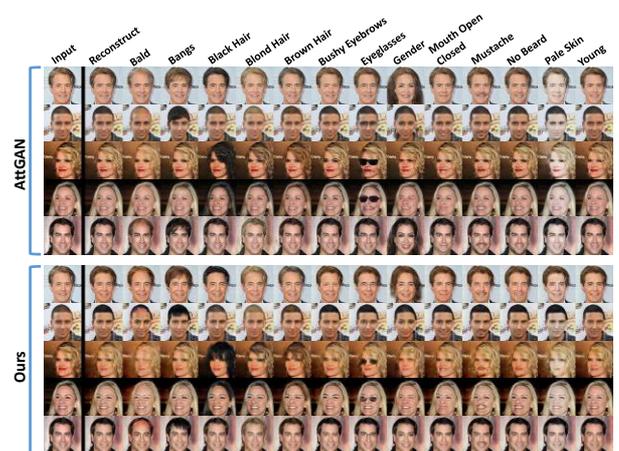

Fig. 5. Modification of attributes in face images. Top: AttGAN. Bottom: Proposed approach. The input images in the first column are transformed one-by-one to all of the corresponding required attributes.

REFERENCES

[1] H. Li, Z. Lin, X. Shen, J. Brandt, G. Hua, "A convolutional neural network cascade for face detection", Proc. IEEE Conf. Comput. Vis. Pattern Recognit. (CVPR), pp. 5325-5334, Jun. 2015.
[2] Y. Bai, Y. Zhang, M. Ding, B. Ghanem, "Finding tiny faces in the wild with generative adversarial network," Proc. IEEE/CVF Conf. Comput. Vis. Pattern Recognit., pp. 21-30, Jun. 2018.
[3] J. Deng, S. Cheng, N. Xue, Y. Zhou, S. Zafeiriou, "UV-GAN: Adversarial facial UV map completion for pose-invariant face recognition," Proc. IEEE Conf. Comput. Vis. Pattern Recognit., pp. 7093-7102, Jun. 2018.
[4] T. Ahonen, A. Hadid, M. Pietikainen, "Face description with local binary patterns: Application to face recognition," IEEE Trans. Pattern Anal. Mach. Intell., vol. 28, pp. 2037-2041, Dec. 2006.
[5] Zhou, Shuchang, et al. "Genegan: Learning object transfiguration and attribute subspace from unpaired data." arXiv preprint arXiv:1705.04932, 2017.
[6] Z. He, W. Zuo, M. Kan, S. Shan, and X. Chen, "Attgan: Facial attribute editing by only changing what you want." IEEE Transactions on Image Processing, Vol. 28, No. 11, Nov. 2019.
[7] Wu, Po-Wei, et al. "Relgan: Multi-domain image-to-image translation via relative attributes." Proceedings of the IEEE International Conference on Computer Vision (ICCV), Seoul, Korea, pp.5914-5922, Oct. 2019.
[8] E. Reinhard, M. Ashikhmin, B. Gooch, and P. Shirley. Color transfer between images.IEEE Comput. Graph., 21:34–41,2001.
[9] C. Barnes, E. Shechtman, A. Finkelstein, and D. Goldman. PatchMatch: A randomized correspondence algorithm for structural image editing. ACM Trans. Graph., 28(3):24, 2009.
[10] M. Alexa, D. Cohen-Or, and D. Levin. As-rigid-as-possible shape interpolation. InSIGGRAPH, 2000.
[11] T. Hassner, S. Harel, E. Paz, and R. Enbar. Effective face frontalization in unconstrained images. In CVPR, 2015.
[12] D. Guo and T. Sim. Digital face makeup by example. In CVPR, 2009.
[13] L. Liu, H. Xu, J. Xing, S. Liu, X. Zhou, and S. Yan. Wow! You are so beautiful today! In ACMMM, 2013.
[14] F. Yang, J. Wang, E. Shechtman, L. Bourdev, and D. Metaxas. Expression flow for 3D-aware face component transfer. In SIGGRAPH, 2011.
[15] I. Kemelmacher-Shlizerman, S. Suwajanakorn, and S. M.Seitz. Illumination-aware age progression. In CVPR, 2014.
[16] Goodfellow, Ian, et al. "Generative adversarial nets." Advances in neural information processing systems, Montreal, Canada, pp. 2672-2680, Dec, 2014.
[17] D. P. Kingma and M. Welling, "Auto-encoding variational bayes," in 2nd International Conference on Learning Representations, - Conference Track Proceedings, Banff, Canada, pp. 1-14, April 2014.
[18] Z. He, W. Zuo, M. Kan, S. Shan, and X. Chen, "Attgan: Facial attribute editing by only changing what you want." IEEE Transactions on Image Processing, Vol. 28, No. 11, Nov. 2019.
[19] J. Kossaifi, L. Tran, Y. Panagakis, and M. Pantic, "Gagan: Geometry-aware generative adversarial networks." Proceedings of the IEEE Conference on Computer Vision and Pattern Recognition(CVPR), Salt Lake City, Utah, pp. 878-887, June. 2018.
[20] Park, Dong-Hun, Vijay Kakani, and Hak-Il Kim. "Automatic Radial Un-distortion using Conditional Generative Adversarial Network." Journal of Institute of Control, Robotics and Systems vol. 25, no. 11, pp. 1007-1013, Nov, 2019.
[21] Perarnau, G., Van De Weijer, J., Raducanu, B., and Alvarez, J. M. "Invertible conditional gans for image editing." arXiv preprint arXiv:1611.06355, 2016.
[22] M. Mirza and S. Osindero, "Conditional generative adversarial nets." arXiv preprint arXiv:1411.1784, Nov. 2014.
[23] T. Kim, B. Kim, M. Cha, and J. Kim, "Unsupervised visual attribute transfer with reconfigurable generative adversarial networks." arXiv preprint arXiv:1707.09798, Jul. 2017.
[24] Liu, Ziwei, et al. "Deep learning face attributes in the wild." Proceedings of the IEEE international conference on computer vision(ICCV), Santiago, Chile pp. 3730-3738, Dec. 2015.